\documentclass[letterpaper, 10 pt, conference]{ieeeconf}  % Comment this line out if you need a4paper

\IEEEoverridecommandlockouts                              % This command is only needed if 
\usepackage{amsmath} % assumes amsmath package installed

\usepackage{graphicx}
\usepackage{cite}
\usepackage[ruled,norelsize]{algorithm2e}
\usepackage{hyperref}   % for hyperlinks in references
\usepackage{caption}
\usepackage{subcaption}
\usepackage{xcolor}
\usepackage{gensymb}  % for degree symbol
\usepackage{makecell} % for inserting line breaks in table cells
\makeatletter
\newcommand{\removelatexerror}{\let\@latex@error\@gobble}
\makeatother

\title{\LARGE \bf
Modeling Evacuee Behavior for Robot-Guided Emergency Evacuation
}

\author{Mollik Nayyar$^{1}$ and Alan R. Wagner$^{2}$ % <-this % stops a space
\thanks{This material is based upon work supported by the National Science Foundation under Grant Number CNS-1830390 and IIS-2045146. Any opinions, findings, and conclusions or recommendations expressed in this material are those of the authors and do not necessarily reflect the views of the National Science Foundation.}% <-this % stops a space
\thanks{$^{1}$Mollik Nayyar is a graduate student in aerospace engineering at the Robot Ethics and Aerial Vehicles lab, The Pennsylvania State University
        {\tt\small mxn244@psu.edu}}%
\thanks{$^{4}$Alan R. Wagner is an assistant professor in the Department of Aerospace Engineering at The Pennsylvania State University
        {\tt\small alan.r.wagner@psu.edu}}%
}

\begin{document}

\maketitle
\thispagestyle{empty}
\pagestyle{empty}

%%%%%%%%%%%%%%%%%%%%%%%%%%%%%%%%%%%%%%%%%%%%%%%%%%%%%%%%%%%%%%%%%%%%%%%%%%%%%%%%
\begin{abstract}
%%%%%% 
This paper considers the problem of developing suitable behavior models of human evacuees during a robot-guided emergency evacuation. We describe our recent research developing behavior models of evacuees and potential future uses of these models. This paper considers how behavior models can contribute to the development and design of emergency evacuation simulations in order to improve social navigation during an evacuation.   
\end{abstract}

%%%%%%%%%%%%%%%%%%%%%%%%%%%%%%%%%%%%%%%%%%%%%%%%%%%%%%%%%%%%%%%%%%%%%%%%%%%%%%%%
\section{INTRODUCTION}

\noindent Our research is attempting to develop robots that are capable of helping people evacuate during an emergency \cite{wagner2021robot}. In order to accomplish this task the robot must not only navigate social environments such as hallways, entry ways and exits, it may also need to avoid collisions or, perhaps, even feint collisions in order get people to move out of its way. Managing the collision avoidance and social navigation problems in such environments is extremely challenging for a variety of reasons \cite{trautman2010unfreezing}. First, emergencies are rare events and difficult to predict. Hence, running real world experiments in which robots attempt to guide evacuees during actual emergencies is not practical. Moreover, emergencies can be unique. Earthquakes can damage buildings making navigation maps unusable. Fires may require evacuation to distant or improvised exits and avoidance of certain areas of a building. Finally, people do not always react rationally during an emergency \cite{leach1994survival, kuligowski2013predicting}. Depending on the type of emergency, humans may not evacuate at all, freeze and remain motionless, or blindly comply with obviously incorrect information. Given the wide range of potential environments, emergencies, robot designs, and evacuees, evaluating the effectiveness of a robot's guidance requires the use of simulation experiments. Although simulation modeling of evacuations \cite{pidd1996simulation,kuligowski2005review,sharma2008multi, santos2004critical} and human behavior modeling during emergencies is an active area of research \cite{leach1994survival, aguirre2005emergency,pan2006computational,pan2007multi}, relatively few groups are exploring the development of robot emergency guides \cite{robinette2016overtrust,boukas2014robot,zhang2015distributed}.

Yet using simulation to evaluate robot designs, behaviors, or evacuation strategies is also problematic for a number of reasons. For instance, the behavior of other evacuees has a strong influence on the experiment's human subject (Fig. \ref{fig:SimEvac})\cite{nayyar2019effective}. In fact, during an evacuation, the behavior of other evacuees is often a determining factor impacting when and how quickly a person evacuates \cite{kuligowski2013predicting}. Our previous simulation experiments demonstrate that people often follow the crowd, in spite of the robot's suggestions \cite{nayyar2019effective}. Moreover, an important aspect of designing a robot that guides people during an emergency is understanding how people will react and respond to the robot \cite{robinette2016assessment}. Yet, simulations may not generate the same visceral response as actual emergencies \cite{robinette2016investigating}. Specifically, will evacuees follow an emergency guide robot? If so, how closely will they follow the robot? Will they follow it through closed doors? Will they hold the door open for the robot? Will they follow the robot if it signals a change in direction? Clearly we have an ethical obligation to thoroughly evaluate emergency evacuation robots prior to their deployment \cite{wagner2020principles}.   

Our research is currently working to address these problems. We are currently in the process of developing behavior models of evacuees that we will then use to create more realistic simulation environments (discussed more below). We are also developing realistic simulations in virtual reality to improve the realism of the experience for the purpose of better understanding social navigation during an emergency. The remainder of this paper presents our preliminary data and ongoing research on this topic. 

\section{MODELING EVACUEE BEHAVIOR}
\noindent This section details our method for creating and using behavior models of evacuees. 

\subsection{Creating Behavior Model}
\begin{figure*}
\centering
\includegraphics[width=0.7\linewidth]{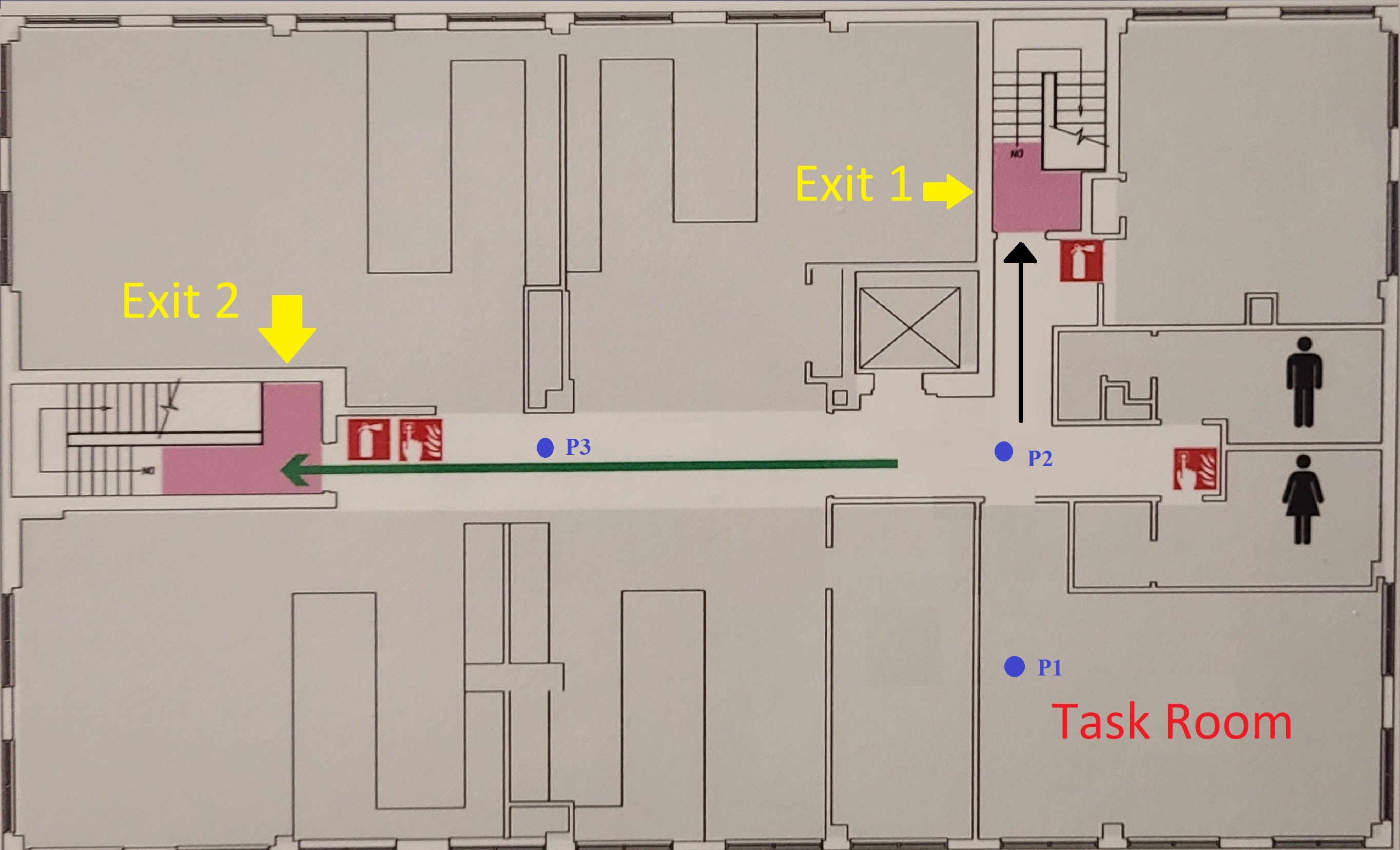} 
\caption{The subject enters the experiment from Exit 1. Upon entering they are introduced to the robot near Exit 1. The robot leads the person to point P1. The subject sits at a cubicle in the Task Room. If the robot system under test is a shepherding system, when the alarm sounds the robot leads the person from point P1 through point P2 to Exit 2 stopping at point P3. If the robot system under test is the handoff system, three individual robots are placed at points P1, P2, and P3 again directing the subject to Exit 2. }
\label{fig:floor_plan}
\end{figure*}

\noindent Given that evacuee behavior is strongly influence by the other people experiencing the emergency, in order to begin to improve the accuracy of an emergency simulation one must also simulate the behavior of the other evacuees \cite{santos2004critical}. To do this we are currently running human subject experiments that capture and record an evacuee's behavior to a robot's guidance directions when an alarm has sounded in a physical environment. Specifically, we have naive human subjects arrive at the experimental site (Fig. \ref{fig:floor_plan}). These subjects are briefly introduced to a guidance robot which leads them to a cubicle in order to read and comment on an article. While reading the article a smoke alarm is sounded. The subject has not been informed that a smoke alarm will go off during the experiment. The subject then chooses whether to follow the robot's guidance directions to an unfamiliar exit or may exit the way they came. 

\begin{figure*}
\centering
\includegraphics[width=0.7\linewidth]{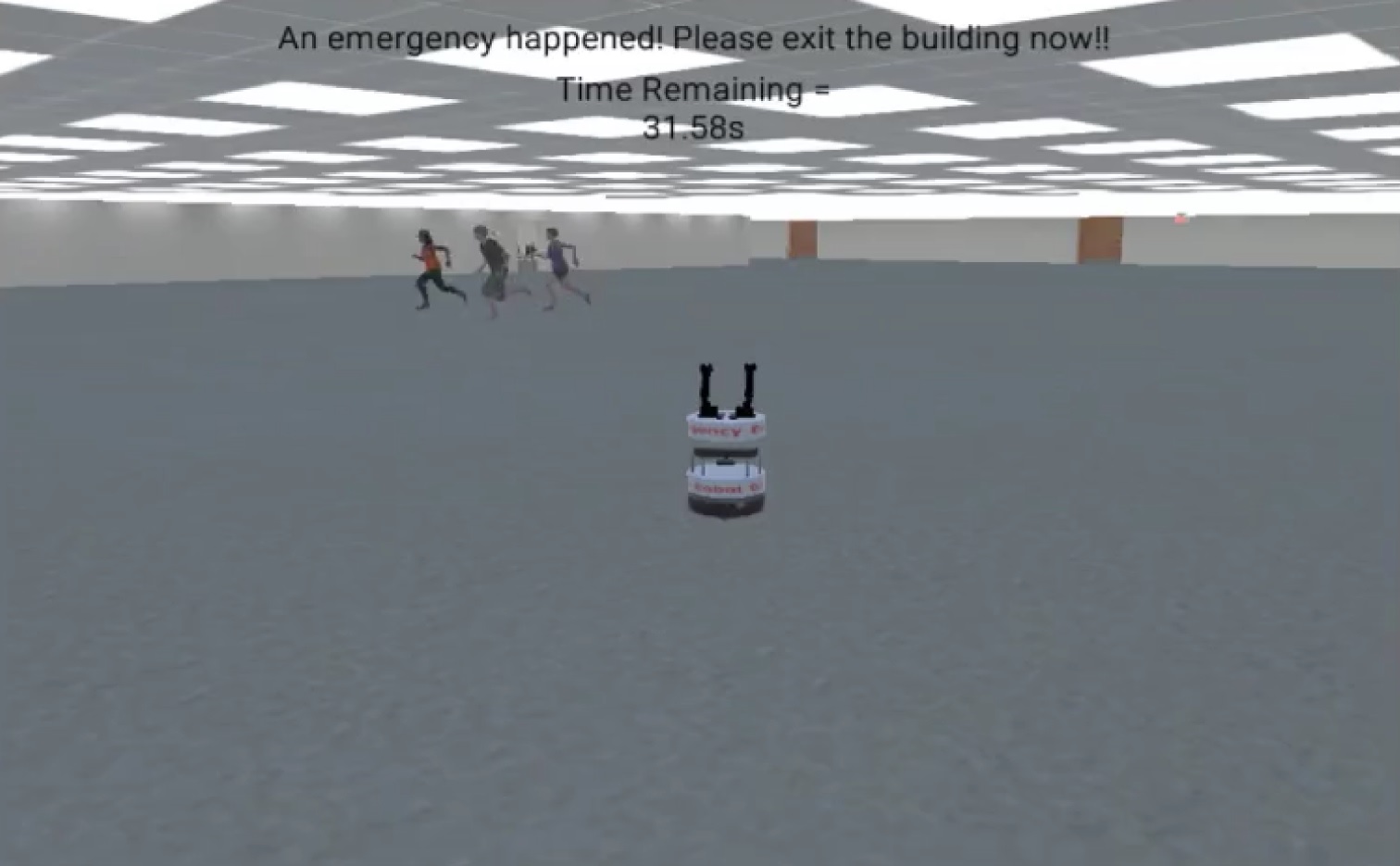} 
\caption{A simulation experiment is depicted in which the subject chooses to follow the robot during an emergency or groups of people running towards a different exit. Our work shows that most people follow the crowd. }
\label{fig:SimEvac}
\end{figure*}

\begin{figure*}
\centering
\includegraphics[width=0.7\linewidth]{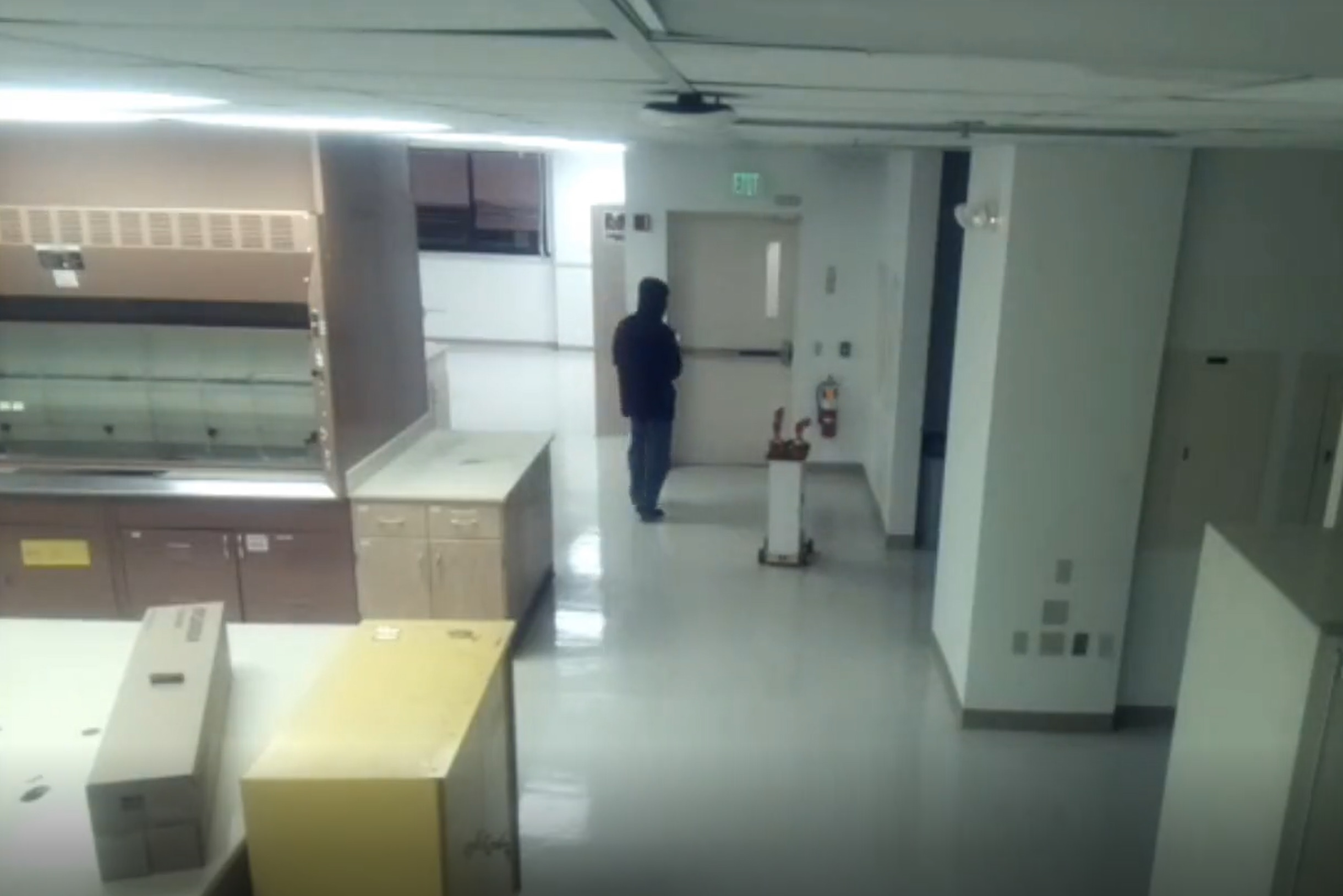} 
\caption{A subject following the robot to an exit. }
\label{fig:SubEvac}
\end{figure*}
We are using this general experimental paradigm to test two different types of emergency evacuation robot systems. The first is a shepherding robot which is a single robot that guides individuals or groups to exits. The second system is a handoff system in which robots are deployed to the important evacuation decision points (e.g. hall intersections). Each robot remains at that position and points to either the next robot along an evacuation path or to the desired exit. These two types of robot systems are being tested with either single evacuees or groups of evacuees. We are recording video data capturing the person's movement through the environment. Soon we will explore some variations of a simple evacuation. Specifically, we intend to capture the subject's behavior when the robot's guidance directions change mid-evacuation, when the robot must guide through a closed door, when confederates ignore the robot's directions, and when the robot instructs the subject to shelter in place.

Nearly seventy human subjects have participated in our experiment to date (Fig. \ref{fig:SubEvac}). Importantly, our intent is to capture the descriptive statistics representing the human subject's behavior. This is in contrast to traditional human-robot interaction (HRI) studies which tend to focus on hypothesis testing of a limited number of hypotheses. These descriptive statistics include measures of the person's speed, local distance from the robot, evacuation time, and the response time of the participant between the sounding of smoke alarm and evacuation action. These measures will allow us to create parameterized behavior models representing each human subject's behavior during the evacuation \cite{vorst2010evacuation}. Although the form of these models has yet to be determined, we envision some form of parameterized finite state machine that can be used in conjunction with a specific set of experimental parameters to generate motion commands representing prototypical evacuees behavior (average motions), exemplar behavior (behavior tied to specific individual subjects), or variations which randomize elements of the subject's behavior.  

\subsection{Using Behavior Models}

\noindent As mentioned, we intend to use these behavior models as a tool to facilitate the evaluation of emergency evacuation robots. In the simulation environment pictured in Figure \ref{fig:SimEvac}, the evacuating crowd runs at a fixed speed to an out of sight exit. While this type of crowd behavior is possible, it is unlikely given that most emergencies generate an initial sense of confusion accompanied with information seeking behaviors \cite{leach1994survival}. 

The most straightforward use of these models is to vary the type of environment (number of exits, number of corridors, etc.) and to use the behavior models to estimate the evacuation time or other evacuation statistics. For example, given a floor plan for school, we might use the behavior models to estimate how long it will take students located in particular classrooms to follow a robot to an exit. These estimates could then be compared to a limited number of physical experiments. 

Another potential way that these models could be used is to create different demographic groups and estimate the amount of time a robot-guided evacuation would require. For example, estimating how long it would take to evacuate able bodied older adults from a nursing home or pregnant mothers could again inform the design and use of the robots (e.g. speed of the robot, number of robots, etc.). Of course, in order to evaluate these different demographic groups we need to study participants of each type. Currently we have several older adults that have participated, but we are not aware of any pregnant women that have joined our study.   

One final method for using these behavior models is to combine different models to generate estimates of evacuation time for different scenarios. For example, using a behavioral model for a shelter in place scenario, followed by a behavior model for an evacuation, followed by a behavior model captured when the robot redirects the person to a different exit. Although the individual behavior models will have been generated by three different subjects, it may be possible to produce a rough estimate of the evacuee's behavior by simply using different models for different events during the evacuation.  

Experimentally, the resulting data could serve as a baseline estimate for in-person experiments or it could govern the behavior of Non-Player Characters (NPCs) in a simulation with a human subject. Hence, the human subject would be surrounded by NPCs using these behavior models with the reactions of the human subject being experimentally observed and recorded. Presented in an immersive virtual reality simulation we believe that we will be able to produce realistic environments with NPC behavior that reflects the behavior of true human subjects.  

\section{LIMITATIONS} \label{limitations}

\noindent We are making a number of important assumptions and there are certainly limitations to our approach. One assumption is that the data generated by one or many evacuees during a robot-guided evacuation will be correlated to and predictive of other evacuees' behavior during different emergencies. This assumption only holds to a limited degree, however. Evacuee behavior after an explosion and fire is unlikely to predict evacuee behavior during small, non-threatening fire. Nevertheless, we believe that our research can provide a type of foothold for representing how people evacuate in the presence of a robot which can then be expanded to new situations and evacuee types. Specifically, by capturing people's behavior towards the robot in response to an alarm, we record information about following speed, likelihood of following, and response to robot guidance actions. This information is necessary for designing evacuation robots and could be useful for related applications, such as path clearing for evacuees \cite{nayyar2021aiding}. 

Furthermore, we recognize that changes to the robot or the environment could influence the resulting models and parameters. It will be necessary to occasionally perform in-person experiments to evaluate if and how much the behavior model has drifted from real world results. Simulation experiments may inform robot and experimental design and focus the research on the most promising hypotheses, yet cannot serve as a complete substitute for in-person experiments.  

Our long-term hope for this work is that a catalog of pedestrian behavior models could be compiled allowing researchers to more accurately simulate the actual social navigation behavior of individuals and groups during emergencies. This catalog could serve as an important tool, perhaps allowing HRI researchers to compare algorithms, systems, and behaviors for a fixed context and reasonably accurate set of behavior models.

%   What behavior model is suitable to model humans in social navigation simulation?
%    How much realism do we need in social navigation simulation?
%    How do we categorize and evaluate context in social navigation?
%    How do we mitigate novelty effect in real-world social navigation experiments?
%   How effective is it to evaluate social navigation models on datasets?

\section{CONCLUSIONS} \label{conclusion}

\noindent This paper has briefly introduced our ongoing research to create behavior models of human evacuees while being guided to an exit by a robot during a simulated emergency. The resulting behavior models will contribute to the realism of our emergency evacuation simulations. This added realism will assist in the design and use of emergency evacuation robots. Over time, we hope to create a catalog of behavior models representing different evacuation scenarios. This catalog could serve as a method to evaluate potential robot social navigation algorithms applied to the emergency evacuation domain.


%%%%%%%%%%%%%%%%%%%%%%%%%%%%%%%%%%%%%%%%%%%%%%%%%%%%%%%%%%%%%%%%%%%%%%%%%%%%%%%%

%%%%%%%%%%%%%%%%%%%%%%%%%%%%%%%%%%%%%%%%%%%%%%%%%%%%%%%%%%%%%%%%%%%%%%%%%%%%%%%%
%\section*{ACKNOWLEDGMENT}

%%%%%%%%%%%%%%%%%%%%%%%%%%%%%%%%%%%%%%%%%%%%%%%%%%%%%%%%%%%%%%%%%%%%%%%%%%%%%%%%

\bibliographystyle{IEEEtran}
\bibliography{IEEEabrv,mybibfile}

\begin{thebibliography}{10}
\providecommand{\url}[1]{#1}
\csname url@rmstyle\endcsname
\providecommand{\newblock}{\relax}
\providecommand{\bibinfo}[2]{#2}
\providecommand\BIBentrySTDinterwordspacing{\spaceskip=0pt\relax}
\providecommand\BIBentryALTinterwordstretchfactor{4}
\providecommand\BIBentryALTinterwordspacing{\spaceskip=\fontdimen2\font plus
\BIBentryALTinterwordstretchfactor\fontdimen3\font minus \fontdimen4\font\relax}
\providecommand\BIBforeignlanguage[2]{{%
\expandafter\ifx\csname l@#1\endcsname\relax
\typeout{** WARNING: IEEEtran.bst: No hyphenation pattern has been}%
\typeout{** loaded for the language `#1'. Using the pattern for}%
\typeout{** the default language instead.}%
\else
\language=\csname l@#1\endcsname
\fi
#2}}

\bibitem{wagner2021robot}
A.~R. Wagner, ``Robot-guided evacuation as a paradigm for human-robot interaction research,'' \emph{Frontiers in Robotics and AI}, vol.~8, 2021.

\bibitem{trautman2010unfreezing}
P.~Trautman and A.~Krause, ``Unfreezing the robot: Navigation in dense, interacting crowds,'' in \emph{2010 IEEE/RSJ International Conference on Intelligent Robots and Systems}.\hskip 1em plus 0.5em minus 0.4em\relax IEEE, 2010, pp. 797--803.

\bibitem{leach1994survival}
J.~Leach, \emph{Survival psychology}.\hskip 1em plus 0.5em minus 0.4em\relax Springer, 1994.

\bibitem{kuligowski2013predicting}
E.~Kuligowski, ``Predicting human behavior during fires,'' \emph{Fire technology}, vol.~49, no.~1, pp. 101--120, 2013.

\bibitem{pidd1996simulation}
M.~Pidd, F.~De~Silva, and R.~Eglese, ``A simulation model for emergency evacuation,'' \emph{European Journal of operational research}, vol.~90, no.~3, pp. 413--419, 1996.

\bibitem{kuligowski2005review}
E.~D. Kuligowski, R.~D. Peacock, B.~L. Hoskins, \emph{et~al.}, \emph{A review of building evacuation models}.\hskip 1em plus 0.5em minus 0.4em\relax US Department of Commerce, National Institute of Standards and Technology~…, 2005.

\bibitem{sharma2008multi}
S.~Sharma, H.~Singh, and A.~Prakash, ``Multi-agent modeling and simulation of human behavior in aircraft evacuations,'' \emph{IEEE Transactions on aerospace and electronic systems}, vol.~44, no.~4, pp. 1477--1488, 2008.

\bibitem{santos2004critical}
G.~Santos and B.~E. Aguirre, ``A critical review of emergency evacuation simulation models,'' 2004.

\bibitem{aguirre2005emergency}
B.~E. Aguirre, ``Emergency evacuations, panic, and social psychology,'' \emph{Psychiatry: Interpersonal and Biological Processes}, vol.~68, no.~2, pp. 121--129, 2005.

\bibitem{pan2006computational}
X.~Pan, \emph{Computational modeling of human and social behaviors for emergency egress analysis}.\hskip 1em plus 0.5em minus 0.4em\relax Stanford University, 2006.

\bibitem{pan2007multi}
X.~Pan, C.~S. Han, K.~Dauber, and K.~H. Law, ``A multi-agent based framework for the simulation of human and social behaviors during emergency evacuations,'' \emph{Ai \& Society}, vol.~22, no.~2, pp. 113--132, 2007.

\bibitem{robinette2016overtrust}
P.~Robinette, W.~Li, R.~Allen, A.~M. Howard, and A.~R. Wagner, ``Overtrust of robots in emergency evacuation scenarios,'' in \emph{2016 11th ACM/IEEE international conference on human-robot interaction (HRI)}.\hskip 1em plus 0.5em minus 0.4em\relax IEEE, 2016, pp. 101--108.

\bibitem{boukas2014robot}
E.~Boukas, I.~Kostavelis, A.~Gasteratos, and G.~C. Sirakoulis, ``Robot guided crowd evacuation,'' \emph{IEEE Transactions on Automation Science and Engineering}, vol.~12, no.~2, pp. 739--751, 2014.

\bibitem{zhang2015distributed}
S.~Zhang and Y.~Guo, ``Distributed multi-robot evacuation incorporating human behavior,'' \emph{Asian Journal of Control}, vol.~17, no.~1, pp. 34--44, 2015.

\bibitem{nayyar2019effective}
M.~Nayyar and A.~R. Wagner, ``Effective robot evacuation strategies in emergencies,'' in \emph{2019 28th IEEE International Conference on Robot and Human Interactive Communication (RO-MAN)}.\hskip 1em plus 0.5em minus 0.4em\relax IEEE, 2019, pp. 1--6.

\bibitem{robinette2016assessment}
P.~Robinette, A.~R. Wagner, and A.~M. Howard, ``Assessment of robot to human instruction conveyance modalities across virtual, remote and physical robot presence,'' in \emph{2016 25th IEEE International Symposium on Robot and Human Interactive Communication (RO-MAN)}.\hskip 1em plus 0.5em minus 0.4em\relax IEEE, 2016, pp. 1044--1050.

\bibitem{robinette2016investigating}
------, ``Investigating human-robot trust in emergency scenarios: methodological lessons learned,'' in \emph{Robust Intelligence and Trust in Autonomous Systems}.\hskip 1em plus 0.5em minus 0.4em\relax Springer, 2016, pp. 143--166.

\bibitem{wagner2020principles}
A.~R. Wagner, ``Principles of evacuation robots,'' in \emph{Living with Robots}.\hskip 1em plus 0.5em minus 0.4em\relax Elsevier, 2020, pp. 153--164.

\bibitem{vorst2010evacuation}
H.~C. Vorst, ``Evacuation models and disaster psychology,'' \emph{Procedia Engineering}, vol.~3, pp. 15--21, 2010.

\bibitem{nayyar2021aiding}
M.~Nayyar and A.~R. Wagner, ``Aiding emergency evacuations using obstacle-aware path clearing,'' in \emph{2021 IEEE International Conference on Advanced Robotics and Its Social Impacts (ARSO)}.\hskip 1em plus 0.5em minus 0.4em\relax IEEE, 2021, pp. 7--14.

\end{thebibliography}

\end{document}